# Intelligent Diagnosis of Alzheimer's Disease Based on Machine Learning

Mingyang Li

School of Liberal Arts and Sciences, China University of Petroleum-Beijing at Karamay, Karamay 834000, Xinjiang, China, limingyang7278@gmail.com

Hongyu Liu

School of Liberal Arts and Sciences, China University of Petroleum-Beijing at Karamay, Karamay 834000, Xinjiang, China, 2021016275@st.cupk.edu.cn

Yixuan Li

School of Liberal Arts and Sciences, China University of Petroleum-Beijing at Karamay, Karamay 834000, Xinjiang, China, 2021016297@st.cupk.edu.cn

Zejun Wang*

School of Liberal Arts and Sciences, China University of Petroleum-Beijing at Karamay, Karamay 834000, Xinjiang, China, wangzj@cupk.edu.cn

Yuan Yuan

State Key Laboratory of Petroleum Resources and Prospecting, China University of Petroleum, Beijing 102249, China, yuany9939@163.com

Honglin Dai

School of Business Administration, China University of Petroleum-Beijing at Karamay, Karamay 834000, Xinjiang, China, honglindai@163.com

Mingyang Li, Hongyu Liu and Yixuan Li These authors contributed equally to this work and should be considered co-first authors.

This study is based on the Alzheimer's Disease Neuroimaging Initiative (ADNI) dataset and aims to explore early detection and disease progression in Alzheimer's disease (AD). We employ innovative data preprocessing strategies, including the use of the random forest algorithm to fill missing data and the handling of outliers and invalid data, thereby fully mining and utilizing these limited data resources. Through Spearman correlation coefficient analysis, we identify some features strongly correlated with AD diagnosis. We build and test three machine learning models using these features: random forest, XGBoost, and support vector machine (SVM). Among them, the XGBoost model performs the best in terms of diagnostic performance, achieving an accuracy of 91%. Overall, this study successfully overcomes the challenge of missing data and provides valuable insights into early detection of Alzheimer's disease, demonstrating its unique research value and practical significance.

**CCS CONCEPTS**

~ Computing methodologies ~ Machine learning ~ Machine learning algorithms

~ Applied computing ~ Life and medical sciences ~ Bioinformatics

~ Computing methodologies ~ Modeling and simulation

**Additional Keywords and Phrases**

Alzheimer's Disease, Missing Data Handling, Machine Learning, XGBoost

# 1 Introduction

Alzheimer's disease (AD) is a progressive neurodegenerative disorder widely acknowledged as a primary cause of dementia [1]. As the disease advances, patients manifest a spectrum of symptoms, including memory impairment, language difficulties, visuospatial skill deficits, and executive function impairments [2]. It is estimated that dementia incurs an annual cost exceeding $60 billion nationwide [3], and the aging populations in both developed and developing nations suggest a doubling of its prevalence within the next 15 years [4]. However, due to limited awareness of AD, most patients are diagnosed in the moderate to severe stages, missing the optimal window for early intervention [5-6]. Thus, there is a pressing need to develop models for Alzheimer's disease identification to enable accurate early diagnosis of AD and Mild Cognitive Impairment (MCI), facilitating timely interventions for maximum cost-effectiveness in future treatments [7]. Although the precise etiology of AD remains elusive, research has established associations between AD and structural brain changes as well as the accumulation of specific substances [8-9].

Currently, scholars have explored various methods for early detection of AD. Gary W. Small has proposed more specific and sensitive early detection through genetic risk measurement and functional brain imaging [10]. Ann D. Cohen and William E. Klunk utilize in vivo amyloid imaging agents and neurodegenerative disease biomarkers to detect early AD pathology and subsequent neurodegenerative disease progression [11]. Helaly, H.A. et al. employ Convolutional Neural Networks (CNN) for early AD detection and medical image classification across different AD stages [12]. To delve deeper into this issue and enhance the quality and efficiency of detection methods, this study leverages machine learning approaches, encompassing the following key steps:

1. Data preprocessing on a large-scale dataset to identify indicators with potential analytical value for AD and subsequently examine their correlation with AD.

2. Construction of multiple diagnostic models using brain structural information and cognitive test results as the dataset. The performance evaluation of these models will shed light on the optimal intelligent diagnostic strategy for AD.

Through these investigations, this paper aims to offer new perspectives and methods for the early diagnosis of Alzheimer's disease and mild cognitive impairment.

# 2 Data Source

The data utilized in this study originates from the Alzheimer's Disease Neuroimaging Initiative (ADNI). The ADNI dataset provides us with a wealth of information for in-depth analysis of early detection and follow-up of Alzheimer's disease (AD). This dataset encompasses information from the following categories of individuals:

1. Cognitively Normal (CN): 4,850 individuals
2. Subjective Memory Complaints (SMC): 1,416 individuals
3. Early Mild Cognitive Impairment (EMCI): 2,968 individuals
4. Late Mild Cognitive Impairment (LMCI): 5,236 individuals
5. Alzheimer's Disease (AD): 1,738 individuals

The data obtained from ADNI is categorized into five main classes based on their characteristics: Basic patient information, Quantitative biomarker values, Cognitive assessments, Quantitative anatomical structure values, Other relevant information. Detailed analysis and utilization of these data will be further explored in subsequent research.

# 3 Data Preprocessing

## 3.1 Handling of Data with Numerous Missing Values

Due to the presence of a substantial amount of missing data in the provided dataset, the following steps were taken for preprocessing:

1. Variables were categorized into five groups based on their type: Personal Information, Gene Expression, PET Scans, MRI Scans, and Cognitive Tests. Within each category, variables with excessively high missing rates were removed.

2. For each specific test administered to individuals, if the test had more than 6 missing data points or lacked essential information, the results of that test were excluded.

Following the treatment of missing values, the following indicators were selected as potential influencing factors in each category:

Personal Information: PTID, GE, PTGENDER, PTEDUCAT, PTRACCAT
Gene Expression: APOE4
PET Scans: FDG, AV45
MRI Scans: WholeBrain, IMAGEUID, Ventricles, ICV
Cognitive Tests: CDRSB, ADAS11, MMSE, ADASQ4, FAQ

## 3.2 Handling of Data with Few Missing Values

For variables with only a small amount of missing data, a machine learning-based approach, namely Random Forest, was employed to predict and impute these missing values. Random Forest is a robust model capable of handling a large number of features and addressing nonlinear relationships. In our application, each variable with missing values was treated as the target variable, while all other complete variables were used as features to train a Random Forest model for predicting missing values. This approach allowed us to obtain a more complete dataset without discarding valuable data.

## 3.3 Handling of Outliers

During data preprocessing, attention was also given to the presence of outliers. Outliers are data points that significantly deviate from other observations and may adversely affect our analysis results. To identify and address these outliers, the interquartile range (IQR) method with box plots was employed. Specifically, the first quartile (Q1) and third quartile (Q3) for each variable were calculated, and the IQR (IQR = Q3 - Q1) was determined. Any values falling below Q1 - 1.5 * IQR or above Q3 + 1.5 * IQR were considered outliers. For these identified outlier data points, replacement with the median was chosen to minimize the overall impact on the data distribution.

## 3.4 Handling of Invalid Data

During the data preprocessing process, it was noted that some variables exhibited overly uniform distributions, providing limited discriminatory information. Therefore, the decision was made to remove these features as follows:

1. For PTRACCAT data, the proportion of "White" individuals exceeded 90% (14,944 out of 16,222). Since the majority of values for this feature were concentrated in a single category, its predictive capacity for the model was likely to be very limited. Consequently, PTRACCAT was removed from the dataset.

2. In PTETHCAT data, it was observed that "Not Hisp/Latino" accounted for 89% (14,443 out of 16,222). Due to the highly uniform distribution of this variable, it was similarly removed from the dataset.

The purpose of these two steps was to reduce the dimensionality of the model while avoiding over-reliance on features lacking discriminative power. By eliminating this invalid data, more focus could be placed on features that may have a significant impact on prediction outcomes.

## 3.5 Data Encoding and Standardization

### 3.5.1 Data Encoding.

After the initial preprocessing of the data, it is imperative to proceed with data encoding. The encoding schema for gender is presented in Table 1, while the encoding for disease progression is delineated in Table 2.

**Table 1:** Gender Encoding

| Male | Female |
|------|--------|
| 1    | 2      |

Table 2: DX_bl Encoding

| CN | SMC | EMCI | LMCI | AD |
|---|---|---|---|---|
| 1 | 2 | 3 | 4 | 5 |

### 3.5.2 Standardization.

For standardizing quantitative data, we use the following formula (1):

$$x_{new} = \frac{x - x_{min}}{x_{max} - x_{min}} \quad (1)$$

Preprocessing the data, given its large volume and numerous variables, a series of steps as shown in Figure 1 are necessary. In the second part, considering that the obtained data is discrete and quantitative, and may not necessarily follow a normal distribution, we employ Spearman correlation coefficient to analyze relationships.

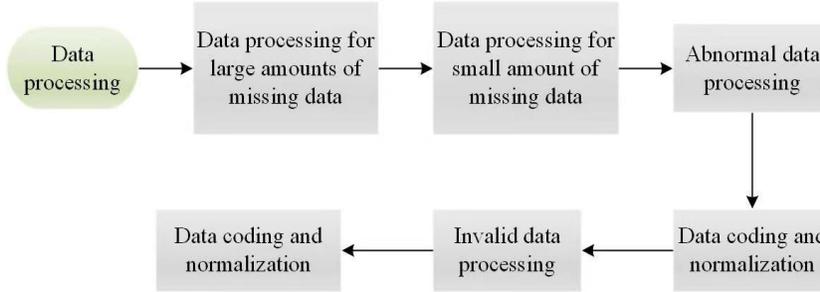

Figure 1: Data Processing Flowchart

# 4 Assumptions

In addressing real-world challenges, we often simplify intricate scenarios through reasonable assumptions, forming and refining models accordingly. The primary assumptions for this analysis are:Assumption 1: The prevalence of Alzheimer's disease is related to gender.

1. Gender and Alzheimer's Prevalence: Epidemiological studies suggest gender influences Alzheimer's prevalence, thus gender is integrated as a potential explanatory variable.

2. Accuracy of Diagnosis and Personal Data: We assume strict adherence to data collection guidelines, ensuring the accuracy of diagnosis dates and personal information.

3. Comprehensiveness of Variables: All variables pertinent to Alzheimer's are detailed in the appendix, streamlining our model's complexity by focusing on the provided variables.

4. Cognitive Tests' Effectiveness: Clinical evidence supports that cognitive tests reliably gauge Alzheimer's progression.

Additional assumptions, if made during model development and analysis, will be elaborated upon in the pertinent segments.

# 5 Correlation Analysis

## 5.1 Spearman Correlation Analysis

After completing the data preprocessing steps, our next task is to explore various factors that may influence the diagnosis of Alzheimer's disease (AD). To investigate the correlation between various data features and the diagnosis of Alzheimer's disease, we have chosen the Spearman correlation coefficient as our analytical tool.

Given that the data we are dealing with is discrete, quantitative, and may not necessarily follow a normal distribution, we have opted for the Spearman correlation coefficient rather than the Pearson correlation coefficient for correlation analysis. The formula for calculating the Spearman correlation coefficient is as follows:

$$\rho = 1 - \frac{6\sum d_i^2}{n(n^2-1)} \quad (2)$$

Where $d_i$ represents the differences in ranks, and n represents the number of data points.

Through calculations, we obtain Spearman correlation coefficients between each feature variable and the diagnosis of Alzheimer's disease and create the following correlation heatmap, as shown in Figure 2.

Generally, the larger the correlation coefficient between two sets of data, the stronger the correlation. We categorize correlations as positive or negative and rank them by the magnitude of the correlation, resulting in the following bar chart:

Based on the Figure 3, we can draw some conclusions. Variables strongly correlated (correlation coefficient absolute value greater than 0.45) with the diagnosis of Alzheimer's disease are CDRSB, FAQ, and ADASQ4. Variables weakly correlated (correlation coefficient absolute value less than 0.1) with the diagnosis of Alzheimer's disease are AGE, ICV, WholeBrain, and PTGENDER.

These results provide important clues for our further modeling and analysis. We will focus on those features strongly correlated with the diagnosis of Alzheimer's disease, aiming to achieve better predictive performance in model construction.

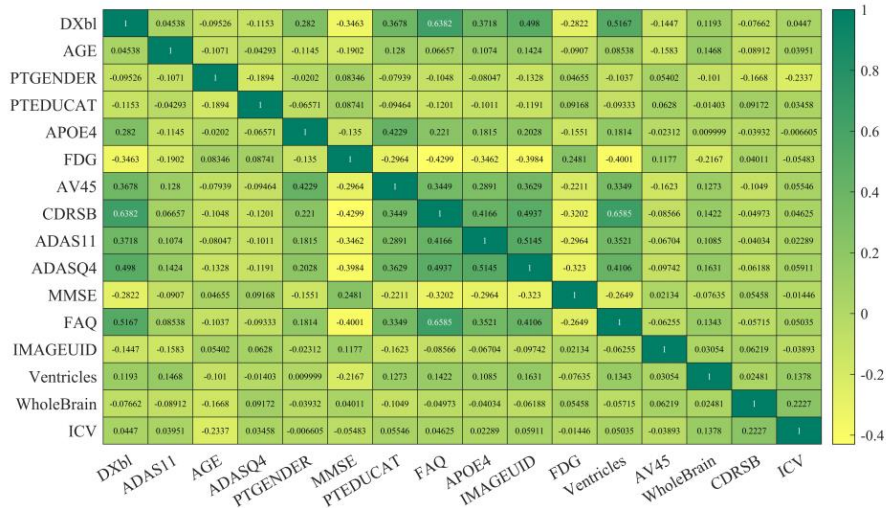

**Figure 2:** Correlation Heatmap

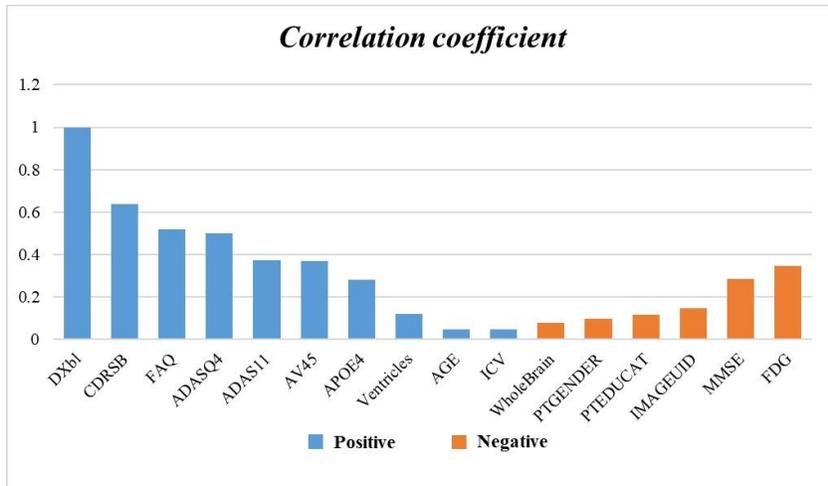

**Figure 3:** Correlation of DX_bl with Other Variables

# 6  Establishment of the Diagnostic Model

## 6.1  Indicator Selection

Based on the results of the data correlation analysis, it is clear that we need to design an intelligent diagnosis for Alzheimer's disease (AD) by combining structural brain features and cognitive-behavioral features. Therefore, we only consider the following features related to Alzheimer's disease for analysis:

MRI measurements: 1.WholeBrain 2.IMAGEUID 3.Ventricles 4.ICV.

Cognitive tests: 1.CDRSB 2. ADAS11 3. MMSE 4.ADASQ4 5.FAQ.

## 6.2  Random Forest-Based Intelligent Diagnosis

Random Forest is an algorithm that employs the concept of ensemble learning by combining multiple decision trees. In machine learning, a random forest is a classifier that consists of multiple decision trees, with the output determined by the majority class output by individual trees.

### 6.2.1  Introduction to Bagging, the Ensemble Learning Algorithm of Random Forest.

The training algorithm of the random forest applies the general technique of Bagging to tree learning. Random Forest is an evolution of Bagging. Random Forest adds random selection of features on top of Bagging, but its basic idea remains within the scope of Bagging. Given a training set $X = X_1, X_2, \ldots, X_n$ and a target set $Y = y_1, y_2, \ldots, y_n$. The random forest method repeats B times, sampling with replacement from the training set, and then training a tree model on these samples ($b = 1,2,\ldots,B$):

1. Sample with replacement n samples from the training set, resulting in $X_b, Y_b$.
2. Train a classification tree or regression tree, $f_b$, based on $X_b, Y_b$.

After training, the prediction for an unknown sample is obtained by averaging the predictions of all the individual regression trees.

The Bagging approach leads to improved performance because it reduces variance while increasing bias. This means that even if the predictions of individual tree models are very sensitive to noise in the training set, it won't be so sensitive for an ensemble of these trees as long as they are uncorrelated.

The ensemble learning model is illustrated as Figure 4:

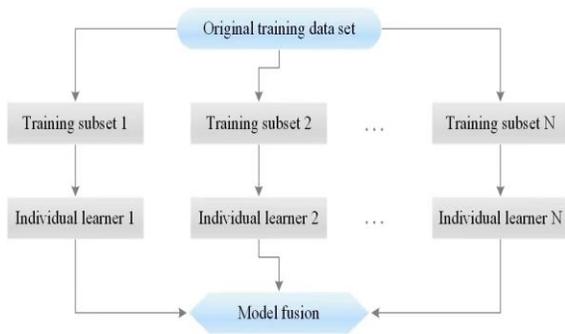

**Figure 4:** Ensemble Learning Model

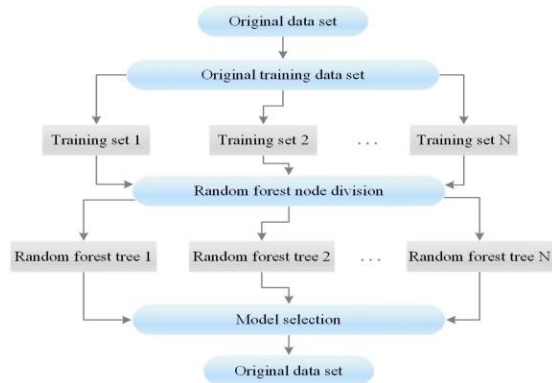

**Figure 5:** Random Forest Model Workflow

### 6.2.2 Construction of Decision Tree Models.

A decision tree is a tree-like predictive model mainly used for instance classification and is a common machine learning method. Decision trees are constructed and used to assist in decision-making, and they represent a special type of tree structure. In random forests, multiple decision trees are built, and the predictions of multiple decision trees are evaluated together to obtain the final result.

In this problem, we choose to use the CART (Classification And Regression Tree) decision tree algorithm, which uses the Gini index to select attributes for splitting. The Gini index can be used to measure the purity of a dataset. The formula for calculating the Gini index is as follows:

$$Gini(p) = 1 - \sum_{j=1}^{m} p_j^2 \qquad (3)$$

Where $p$ is the distribution of samples, $p_j$ is the probability of the j-th class of samples, and m is the total number of classes.

The smaller the Gini index value, the higher the purity of the dataset. When constructing a decision tree, we typically choose the attribute with the smallest Gini index as the splitting attribute.

Using this approach, we can build a decision tree model with good classification performance, serving as the basis for the random forest algorithm.

### 6.2.3 Establishment and Testing of the Random Forest Model.

Step 1: Remove m samples from the original dataset N, and repeat this process n times to create a new sample set as the training set.

Step 2: Train a decision tree model using each of the n training sets.

Step 3: During the construction of each decision tree, calculate the optimal split node based on the characteristics of the attribute data and build an optimal decision tree.

Step 4: The multiple decision trees generated form a random forest. During the prediction process, multiple random forests collectively determine the prediction of missing values.

The model workflow is illustrated in Figure 5.

The training set constitutes 70% of the data, and the test set makes up 30%. The testing was performed 10 times, and the accuracy of the training and test sets is illustrated in Figure 6.

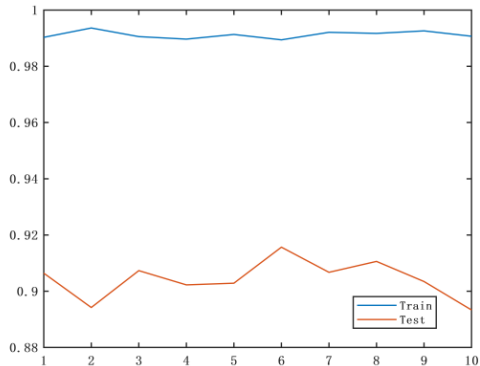

**Figure 6:** Random Forest Model Test Results

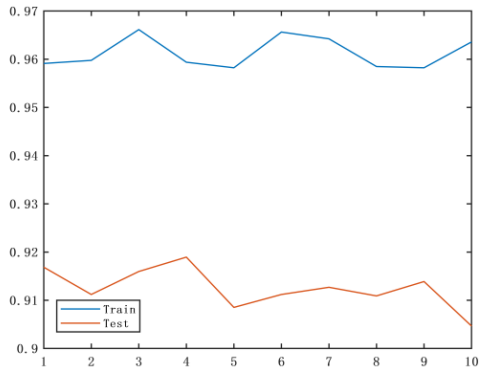

**Figure 7:** XGBoost Model Test Results

The above graph shows that the accuracy of the test set fluctuates around 0.9, while the accuracy of the training set stabilizes at around 0.98. Therefore, the random forest model is practical for addressing the intelligent diagnosis of Alzheimer's disease, achieving an overall accuracy of over 90%.

## 6.3  Establishment and Testing of the XGBoost Model

XGBoost is an ensemble of classification and regression trees. Typically, the strength of a single tree is insufficient. Ensemble models combine predictions from multiple trees and accumulate the prediction scores of each tree to obtain the final score. Unlike traditional GBDT, XGBoost performs a second-order Taylor expansion on the loss function and introduces a regularization term. This can be used to control model complexity, accelerate convergence, reduce the risk of overfitting, and ensure model robustness.

The training set constitutes 70% of the data, and the test set makes up 30%. The testing was performed 10 times, and the accuracy of the training and test sets is illustrated in Figure 7.

The graph above shows that the accuracy of the test set fluctuates slightly around 0.91, while the accuracy of the training set fluctuates slightly around 0.96. Therefore, XGBoost is effective for addressing the intelligent diagnosis of Alzheimer's disease.

## 6.4   Model Construction and Testing for SVM

Support Vector Machine (SVM) is a machine learning algorithm that uses the "margin" as the loss function. The main objective of the learning process is to satisfy the goal of maximizing the margin, ultimately achieving classification. First, for any hyperplane:

$$\omega^T x + b = 0 \quad\quad (4)$$

Where $\omega$ and $x$ are both vectors.

The training set constitutes 70% of the data, and the test set makes up 30%. The testing was performed, and the accuracy of the training and test sets is illustrated in Figure 8.

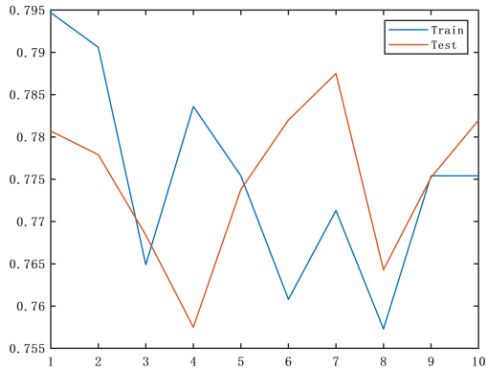

**Figure 8:** SVM Model Test Results

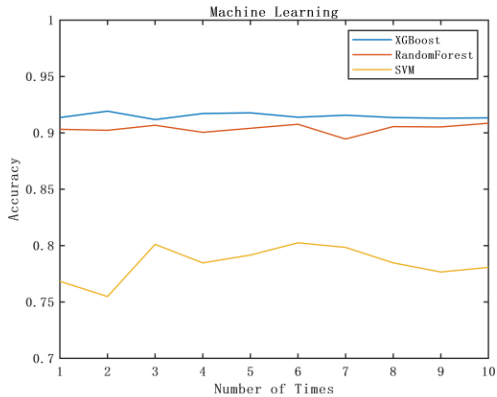

**Figure 9:** Accuracy of the Three Models

The graph above shows that the accuracy of the test set is slightly above 0.76, while the accuracy of the training set stabilizes around 0.75. Therefore, SVM is somewhat insufficient for addressing the intelligent diagnosis of Alzheimer's disease, with an overall accuracy of around 75%.

## 6.5   Model Comparison

For the intelligent diagnosis of Alzheimer's disease in Figure 9, the accuracy of the three models, XGBoost, Random Forest, and SVM, is ranked in descending order. XGBoost has the highest accuracy at 91%, followed by Random Forest at 90%, and SVM at 75%. Therefore, among these three algorithms, XGBoost has the highest accuracy and is the best diagnostic method for Alzheimer's disease.

# 7   Conclusions

In conclusion, this study leveraged the Alzheimer's Disease Neuroimaging Initiative (ADNI) dataset to conduct an in-depth investigation into early detection and disease progression in Alzheimer's disease (AD). Facing the challenge of dealing with a significant amount of missing data, we employed various innovative strategies, including using the random forest algorithm for data imputation, handling outlier data, and filtering out invalid data. Through Spearman rank correlation coefficient analysis, we identified several features strongly correlated with AD diagnosis.

In the phase of machine learning model construction and testing, we explored three models: random forest, XGBoost, and support vector machine (SVM). The results indicated that the XGBoost model achieved the highest diagnostic accuracy, exceeding 90%.

Overall, this study successfully addressed the challenge of missing data and provides valuable insights into early detection in Alzheimer's disease. It not only holds academic significance but also offers strong support for the clinical diagnosis and treatment of Alzheimer's disease.

### ACKNOWLEDGMENTS

This work is supported by Youth Foundation of China University of Petroleum-Beijing at Karamay (No.XQZX20230034).